# The Measure of a Model [*]

**Rebecca Bruce†, Janyce Wiebe‡, Ted Pedersen†**

†Department of Computer Science and Engineering
Southern Methodist University
Dallas, TX 75275-0112

and

‡Department of Computer Science
New Mexico State University
Las Cruces, NM 88003

rbruce@seas.smu.edu, wiebe@cs.nmsu.edu, pedersen@seas.smu.edu

## Abstract

This paper describes measures for evaluating the three determinants of how well a probabilistic classifier performs on a given test set. These determinants are the appropriateness, for the test set, of the results of (1) feature selection, (2) formulation of the parametric form of the model, and (3) parameter estimation. These are part of any model formulation procedure, even if not broken out as separate steps, so the tradeoffs explored in this paper are relevant to a wide variety of methods. The measures are demonstrated in a large experiment, in which they are used to analyze the results of roughly 300 classifiers that perform word-sense disambiguation.

## Introduction

This paper presents techniques that can be used to analyze the formulation of a probabilistic classifier. As part of this presentation, we apply these techniques to the results of a large number of classifiers, developed using the methodology presented in (2), (3), (4), (5), (12) and (16), which tag words according to their meanings (i.e., that perform *word-sense disambiguation*). Other NLP tasks that have been performed using probabilistic classifiers include part-of-speech tagging (11), assignment of semantic classes (8), cue phrase identification (9), prepositional phrase attachment (15), other grammatical disambiguation tasks (6), anaphora resolution (7) and even translation equivalence (1). In fact, it could be argued that any problem with a known set of possible solutions can be cast as a classification problem.

A *probabilistic classifier* assigns, out of a set of possible classes, the one that is most probable according to a probabilistic model. The model expresses the relationships among the *classification variable* (the variable representing the classification tag) and variables that correspond to properties of the ambiguous object and the context in which it occurs (the *non-classification variables*). Each model uniquely defines a classifier.

The basic premise of a probabilistic approach to classification is that the process of assigning object classes is non-deterministic, i.e., there is no infallible indicator of the correct classification. The purpose of a probabilistic model is to characterize the uncertainty in the classification process. The probabilistic model defines, for each class and each ambiguous object, the probability that the object belongs to that class, given the values of the non-classification variables.

The main steps in developing a probabilistic classifier and performing classification on the basis of a probability model are the following.[1]

1. **Feature Selection:** selecting informative contextual features. These are the properties of the ambiguous object and the context in which it occurs that are indicative of its classification. Typically, each feature is represented as a random variable (a non-classification variable) in the probabilistic model. Here we will use $F_i$ to designate a random variable that corresponds to the $i$th contextual feature, and $f_i$ to designate the value of $F_i$. The contextual features play a very important role in the performance of a model. They are the representation of context in the model, and it is on the basis of them that we must distinguish among the classes of objects.

2. **Selection of the parametric form of the model.** The form of the model expresses the joint distribution of all variables as a function of the values of a set of unknown *parameters*. Therefore, the parametric form of a model specifies a **family** of distributions. Each member of that family corresponds to a different set of values for the unknown parameters. The form of a model specifies the stochastic relationships, the interdependencies, that exist among the variables. The parameters define the distributions of the sets of interdependent variables, i.e., the probabilities of the various combinations of the values of the interdependent variables. As an illustration, consider the following three parametric forms, each specifying different sets of interdependencies among variables in describing the joint distribution of a classification variable, $Tag$, and a set of non-classification variables, $F_1$ through $F_n$. In the equations below, $tag$ represents the value of the classification variable and the $f_i$'s denote the values of the non-classification variables.

---

[*]This research was supported by the Office of Naval Research under grant number N00014-95-1-0776.

[1]Although these are always involved in developing probabilistic classifiers, they may not be broken out into three separate steps in a particular method; an example is decision tree induction (14).

The model for interdependence among all variables:

$$\forall\, tag, f_1, f_2, \ldots, f_n \; P(tag, f_1, f_2, \ldots, f_n) = P(tag, f_1, f_2, \ldots, f_n) \quad (1)$$

The model for conditional independence among all non-classification variables given the value of the classification variable:

$$\forall\, tag, f_1, f_2, \ldots, f_n \; P(tag, f_1, f_2, \ldots, f_n) = P(f_1|tag) \times \cdots \times P(f_n|tag) \times P(tag) \quad (2)$$

The model for independence among all variables:

$$\forall\, tag, f_1, f_2, \ldots, f_n \; P(tag, f_1, f_2, \ldots, f_n) = P(tag) \times P(f_1) \times P(f_2) \times \cdots \times P(f_n) \quad (3)$$

The objective in defining the parametric form of a model is to describe the relationships among all variables in terms of only the most important interdependencies. While it is always true that all variables can be treated as interdependent (equation 1), if there are several features, such a model could have too many parameters to estimate in practice. The greater the number of interdependencies expressed in a model the more *complex* the model is said to be.

3. **Estimation of the model parameters from the training data.** While the form of a model identifies the relationships among the variables, the parameters express the uncertainty inherent in those relationships. Recall that the parameters of a model describe the distributions of the sets of interdependent variables by defining the likelihood of seeing each combination of the values of those variables. For example, the parameters of the model for independence are the following:

$$\forall\, tag, f_1, f_2, \ldots, f_n :$$
$$P(tag),\; P(f_1),\; P(f_2),\; \ldots,\; P(f_n)$$

There are no interdependencies in the model for independence, so the parameters describe the distributions of the individual variables.

In the model for conditional independence stated in equation 2, the parameters are as follows:

$$\forall\, tag, f_1, f_2, \ldots, f_n :$$
$$P(f_1|tag),\; \ldots,\; P(f_n|tag),\; P(tag)$$

Each parameter in this model describes the distribution of the tag in combination with a single contextual feature.

The parameters of any model are *estimated* if their values are based on functions of a data sample (i.e., *statistics*) as opposed to properties of the population.

4. **Assessment of the likelihood of each tag:** use of the completed model to compute the probability of assigning each tag to the ambiguous object, given the values of the non-classification variables. This probability function is the following conditional or context-specific distribution of tags, where the $f_i$'s now denote the values assumed by the non-classification variables in the specific context being considered.

$$\forall\, tag \; P(tag|f_1, f_2, f_3, \ldots, f_n) \quad (4)$$

5. **Ambiguity resolution:** assignment, to the ambiguous object, of the tag with the highest probability of having occurred in combination with the known values of the non-classification variables. This assignment is based on the following function (where $\widehat{tag}$ is the value assigned):

$$\widehat{tag} = \overset{argmax}{tag} \; P(tag|f_1, f_2, f_3, \ldots, f_n) \quad (5)$$

In most cases,[2] the process of applying a probabilistic model to classification (i.e., steps (4) and (5) above) is straightforward. The focus of this work is on formulating a probabilistic model (steps (1)-(3)); these steps are crucial to the success of any probabilistic classifier. We describe measures that can be used to evaluate the effect of each of these three steps on classifier performance. Using these measures, we demonstrate that it is possible to analyze the contribution of each step as well as the interdependencies that exist between these steps.

The remainder of this paper is organized as follows. The first section is a description of the experimental setup used for the investigations performed in this paper. Next, the evaluation measures that we propose are presented, followed by a discussion of the results and finally a presentation of our conclusions.

---

[2] When the values of all non-classification variables are known and there are no interdependent ambiguities among the classes.

# The Experimental Setup

In this paper, we analyze the performance of classifiers developed for the disambiguation of twelve different words. For each of these words, we develop a range of classifiers based on models of varying complexity. Our purpose is to study the contribution that each of feature selection, selection of the form of a model, and parameter estimation makes to overall model performance. In this section, we describe the basic experimental setup used in these evaluations, in particular, the protocol used in the disambiguation experiments and the procedure used to formulate each model.

## Protocol for the Disambiguation Experiments

There are three parameters that define a word-sense disambiguation experiment: (1) the choice of words and word meanings (their number and type), (2) the method used to identify the "correct" word meaning, and (3) the choice of text from which the data is taken. In these experiments, the complete set of non-idiomatic senses defined in the Longman's Dictionary of Contemporary English (LDOCE) (13) is used as the tag set for each word to be disambiguated. For each use of a targeted word, the best tag, from among the set of LDOCE sense tags, is determined by a human judge. The tag assigned by the classifier is accepted as correct only when it is identical to the tag pre-selected by the human judge.

All data used in these experiments are taken from the Penn Treebank Wall Street Journal corpus (10). This corpus was selected because of its availability and size. Further, the POS categories assigned in the Penn Treebank corpus are used to resolve syntactic ambiguity so that word-meaning disambiguation occurs only after the syntactic category of a word has been identified.

The following words were selected for disambiguation based on their relatively high frequency of occurrence and the appropriateness of their sense distinctions for the textual domain.

- Nouns: *interest*, *bill*, *concern*, and *drug*.
- Verbs: *close*, *help*, *agree*, and *include*.
- Adjectives: *chief*, *public*, *last*, and *common*.

Because word senses from a particular dictionary are used, the degree of ambiguity for each word is fixed, and the overall level of ambiguity addressed by the experiment is determined by this selection of words. For each of these words, the sense tags and their distributions in the data are presented in Tables 1 through 3.

| Noun senses of *interest*: (total count: 2368) | |
|---|---|
| 1 "readiness to give attention": | 15% |
| 2 "quality of causing attention to be given": | <1% |
| 3 "activity, subject, etc., which one gives time and attention to": | 3% |
| 4 "advantage, advancement, or favor": | 8% |
| 5 "a share in a company, business, etc.": | 21% |
| 6 "money paid for the use of money": | 53% |
| Noun senses of *concern*: (total count: 1488) | |
| 1 "a matter that is of interest or importance": | 3% |
| 2 "serious care or interest": | 2% |
| 3 "worry; anxiety": | 32% |
| 4 "a business; firm": | 64% |
| Noun senses of *bill*: (total count: 1335) | |
| 1 "a plan for a law, written down for the government to consider": | 69% |
| 2 "a list of things bought and their price": | 10% |
| 4 "a piece of paper money" (extended to include treasury bills): | 21% |
| Noun senses of *drug*: (total count: 1217) | |
| 1 "a medicine or material used for making medicines": | 58% |
| 2 "a habit-forming substance": | 42% |

Table 1: Data summary - Nouns.

## Feature Selection

For simplicity, the contextual features used in all models were selected per the following schema. All models developed for each of the 12 words incorporate the following types of contextual features: one morphological feature, three collocation-specific features, and four class-based features, with POS categories serving as the word classes. All models developed for the same word (which are models of varying complexity) contain the same features.

The morphological feature describes only the suffix of the base lexeme of the word to be disambiguated: the presence or absence of the plural form, in the case of the nouns, and the suffix indicating tense, in the case of the verbs; the adjectives have no morphological feature under this definition.

The values of the class-based variables are a set of 25 POS tags derived from the first letter of the tags used in the Penn Treebank corpus. Each model contains four variables representing class-based contextual features: the POS tags of the two words immediately preceding and the two words immediately succeeding the ambiguous word. All variables are confined to sentence boundaries; extension beyond the sentence boundary is indicated by a null POS tag (e.g., when the ambiguous word

| Verb senses of *close*: (total count: 1534) | |
|---|---|
| 1 "to (cause to) shut": | 2% |
| 2 "to (cause to) be not open to the public": | 2% |
| 3 "to (cause to) stop operation": | 20% |
| 4 "to (cause to) end": | 68% |
| 6 "to (cause to) come together by making less space between": | 2% |
| 7 "to close a deal" (extended from an idiomatic usage): | 6% |
| **Verb senses of *agree*: (total count: 1356)** | |
| 1 "to accept an idea, opinion, etc., esp. after unwillingness or argument": | 78% |
| 2 "to have or share the same opinion, feeling, or purpose": | 22% |
| 3 "to be happy together; get on well together": | <1% |
| **Verb senses of *include*: (total count: 1558)** | |
| 1 "to have as a part; contain in addition to other parts": | 91% |
| 2 "to put in with something else - human subject": | 9% |
| **Verb senses of *help*: (total count: 1398)** | |
| 1 "to do part of the work for - human object": | 20% |
| 2 "to encourage, improve, or produce favorable conditions for - inanimate object": | 75% |
| 3 "to make better - human object": | 4% |
| 4 "to avoid; prevent; change - inanimate object": | 1% |

Table 2: Data summary - Verbs.

| Adjective senses of *common*: (total count:1111) | |
|---|---|
| 1 "belonging to or shared equally by 2 or more": | 7% |
| 2 "found or happening often and in many places; usual": | 8% |
| 3 "widely known; general; ordinary": | 3% |
| 4 "of no special quality; ordinary": | 1% |
| 6 "technical, having the same relationship to 2 or more quantities": | <1% |
| 7 "as in the phrase 'common stock'" (not in LDOCE): | 80% |
| **Adjective senses of *last*: (total count: 3180)** | |
| 1 "after all others": | 6% |
| 2 "on the occasion nearest in the past": | 93% |
| 3 "least desirable (not in LDOCE)": | <1% |
| **Adjective senses of *chief*: (total count: 1036)** | |
| 1 "highest in rank": | 86% |
| 2 "most important; main": | 14% |
| **Adjective senses of *public*: (total count: 867)** | |
| 1 "of, to, by, for, or concerning people in general": | 56% |
| 2 "for the use of everyone; not private": | 8% |
| 3 "in the sight or hearing of many people; not secret or private": | 11% |
| 4 "known to all or to many": | 3% |
| 5 "connected or concerned with the affairs of the people, esp. with government": | 16% |
| 6 "(of a company) to become a public company" (extended from an idiomatic usage): | 6% |
| 7 "as in the phrase 'public TV' or 'public radio'" (not in LDOCE): | 1% |

Table 3: Data summary - Adjectives.

appears at the start of the sentence, the POS tags to the left have the value null).

Three collocation-specific variables are included in each model, where the term *collocation* is used loosely to refer to a specific spelling form occurring in the same sentence as the ambiguous word. While collocation-specific variables are, by definition, specific to the word being disambiguated, the procedure used to select them is general. The search for collocation-specific variables is limited to the 400 most frequent spelling forms in a data sample composed of sentences containing the ambiguous word. Out of those 400, the three spelling forms whose presence was found to be the most dependent on the value of the classification variable, using the test for independence described in (12), were selected as the collocational variables for that word.

### Formulation of a Range of Parametric Forms

To support these experiments, for each word, a range of models of varying complexity were formulated, with each model defining a new classifier. To distinguish among these models, we introduce a measure of model complexity: the total number of pairwise interdependencies that are specified in the model. For each word, the model of maximal complexity is the model in which all variables are considered to be interdependent (equation 1). The model of minimal complexity formulated for each word is the model in which all non-classification variables are considered to be conditionally independent given the value of the classification variable (equation 2); this is the simplest model that still uses each non-classification variable in predicting the value of the classification variable.

The formulation of these models is conducted as a series of stepwise refinements, starting with the model of maximal complexity. At each step, a new model is formulated from the current model as follows (initially the current model is the starting model). Each of the pairwise interdependencies in the current model is evaluated, using a goodness-of-fit test. The test used is an exact test (12) for evaluating the interdependency between two variables, where two variables are interdependent if they are not conditionally (or fully) independent. The test determines the degree to which that interdependency is manifested in the training data. The new (less complex) model formulated is the current model with the interdependency that is least apparent in the training data removed. The new model is used to classify the test data and then serves as the current model in the next simplification step. A more complete description of this procedure can be found in (2).

## Parameter Estimation

In these experiments, we use *maximum-likelihood estimates* (*M.L. estimates*) of the model parameters. The theoretical motivation behind this approach is intuitively appealing: the model parameters are represented by the numerical values that maximize the probability of generating the training data from a model of the specified form. The implementation is straightforward. For each set of interdependent variables in the model, the associated parameters are the probabilities of the combinations of the values of those variables. The estimates of those parameters are equal to the relative frequencies with which those combinations occur in the training data. The drawback is that the estimates of parameters corresponding to events that occur infrequently in the training data are not reliable; for example, if an event is not observed in the training data, then the estimated probability of that event is zero.

## Description of Evaluation Measures

This paper describes measures that can be used to examine the appropriateness, for the test set, of the features used in a model, the parametric form of the model, and the parameter estimates. Figures 1-12 plot model complexity against a number of model performance measures. The gaps between the overall classification performance of a model (indicated as "Overall Model" in the figures) and the other measures is variously due to error introduced by the three factors under study. We first define all of the performance measures shown in the figures, and then discuss what can be concluded from the relationships among measures.

Below, a *completed model* is a model in which the features have been specified; the parametric form has been specified; and the parameters have been estimated.

1. **Overall Model Performance.** Given a completed model in which the parameters have been estimated from the training data:

the *overall model performance* is the percentage of the test set tagged correctly by a classifier using that model to tag the test set.

**Comments:** Other widely-used loss functions are entropy, cross-entropy, and squared error.

2. **Lower Bound.** Let $FT$ be the most frequently-occurring (correct) tag for a word in the test set. The *lower bound* for that word is the percentage of the test set assigned tag $FT$.

**Comments:** The classification performance of a probabilistic model should not be worse than that of the simplest model, the model for independence:

$$\forall\, tag, f_1, f_2, \ldots, f_n\ P(tag, f_1, f_2, \ldots, f_n) = $$
$$P(tag) \times P(f_1) \times P(f_2) \times \cdots \times P(f_n) \quad (6)$$

Because the probability of seeing each value of the classification variable (i.e., each tag) is independent of the context, this model assigns every object the most frequently occurring tag:

$$\widehat{tag} = \overset{argmax}{tag}\ P(tag|f_1, f_2, f_3, ..., f_n) = $$
$$\overset{argmax}{tag}\ P(tag) \quad (7)$$

Therefore, the proportion of the test set belonging to the most frequently occurring tag establishes the lower bound on model performance. For example, if 60% of the instances of the target word in the test set have the same sense, say sense 1, then the lower bound for model performance is 60%.

3. **Recall.** Given a completed model in which the parameters have been estimated from the training data:

*Recall* is the percentage of the test set that is assigned some tag (correct or not) by a classifier using that model to tag the test set.

**Comments:** An ambiguous word in the test set is not assigned a tag when the parameter estimates characterizing its context are zero. Because M.L. parameter estimates are used, all combinations of variable values that are not observed in the training data are not expected to occur (have zero probability).

The percentage of the test set that is assigned a tag corresponds to the percentage of the combinations of variable values observed in the test set that were also observed in the training data.

4. **Precision.** Given a completed model in

which the parameters have been estimated from the training data:

Of the portion of the test set that is assigned some tag by a classifier using that model to tag the test set, *precision* is the percentage that is tagged correctly.

**Comments:** Equivalently, this measure is:

$$1 - (recall - overallModelPerformance) \quad (8)$$

We will use the term *misclassification error* for $1 - precision$, which is the gap between recall and overall model performance.

**5. Appropriateness of the Parametric Form for the Test Set** (or, the **Measure of Form**). This measure is computed to be identical to the *overall model performance*, except that the parameters are estimated from the **test** data, rather than the training data. That is, given a completed model in which the parameters have been estimated from the **test** data:

The *appropriateness of the parametric form for the test set* is the percentage of the test set tagged correctly by a classifier using that model to tag the test set.

**Comments:** Because the model is trained and tested on the same data, the parameter estimates are optimal for that data. Thus, variation of this performance measure is due only to differences in the parametric form of the model.

**6. Appropriateness of the Feature Set for the Test Set** (or, the **Measure of Feature-Set**). This is equal to the measure of form of the maximally-complex model (i.e., the model that includes all possible interdependencies).

**Comments:** Recall that the measure of form involves a model that is both trained and tested on the test set. When the model is maximally complex and the parameters are estimated from the same data that is being tagged, the model describes the exact joint distribution apparent in that data. Suppose that, for each combination of the values of the non-classification variables that occurs in the test set, the tag is the same for all occurrences (and is the correct one). Then, the features are perfect for the test set: each combination of non-classification variables that occurs in the test data uniquely determines the correct tag. In this case, the performance of the full model is necessarily 100%.

If the performance is not 100%, since the model describes the exact joint distribution, the degraded performance can only be due to the lack of complete discriminatory power of the features—i.e., there are combinations of feature values with which more than one tag occurs. The incorrect answers are the less frequent tags in contexts where there are multiple tags (see equation 4).

Consider the gap between recall and overall model performance, i.e., misclassification error. This gap is the percentage of the objects tagged that were tagged incorrectly. The incorrectness is due to some combination of (1) the features being imperfect, (2) the form being inadequate, and (3) the parameter estimates being inappropriate. In the remainder of this paper, we will analyze the contribution of each of these three factors, using the performance measures defined above.

## Results

In Figures 1 through 12 we use the measures described above to analyze the performance of a series of models for each of the 12 words listed in the section on experimental setup. For each word, we formulate a range of models of varying complexity. The model of maximal complexity is the model in which all variables are considered to be interdependent (equation 1). The model of minimal complexity that is formulated is the model in which all non-classification variables are considered to be conditionally independent given the value of the classification variable (equation 2).

For each word to be disambiguated there is a figure depicting the various measures of model performance as a function of model complexity, where model complexity ranges from the maximal to the minimal model.

Our purpose is to study the effect that each of the three facets of model formulation has on model performance. By evaluating each facet independently we can gauge the impact that each has on the overall performance of a classifier. This is important for many reasons, but here our primary concern is understanding the limitations of model performance.

Using the measures described previously, we are able to demonstrate four main points regarding model formulation. Note that all measures used in establishing these claims are applied with respect to some specific test set and therefore the results are dependent on the characteristics of the particular test set being used.

### The feature set fixes the upper bound of model performance.

As discussed in item 6, if the feature set is ideal for the test set, then each context will uniquely correspond to a single tag. In other words, the feature set is an infallible indicator of the correct tag. When this is not the case (i.e., there are contexts in which two or more tags occur), then all but the most frequently occurring tag (for that context) will be misclassified, and there is nothing that can be changed with regards to the parametric form or the parameter estimates to remedy this situation. Therefore the feature set establishes the upper bound of model performance. This is demonstrated in Figures 1 through 12. It is interesting to note that for four of the words ("bill", "chief", "include", and "concern") the feature set was optimal for the test set (i.e., the measure of feature-set was 100%). Even in the worst case, the error introduced by the lack of discriminatory power of the feature set did not exceed 8%. Note that when the feature set is not optimal, the resulting error affects the precision of the model. This can be observed by comparing the gap between recall and overall model performance (the misclassification error, equivalent to $1 - precision$) for models with relatively large feature-related error (such as the models for "public") to that of models in which the features are optimal, such as those for "bill" and "include". When the feature set is optimal, it contributes nothing to misclassification error. When this is the case, misclassification error is strictly a function of the appropriateness of the parametric form and the parameter estimates. We consider measures of these contributions next.

### As the complexity of the model is reduced, important information is lost from the parametric form.

The measure of the appropriateness of the parametric form (the measure of form) is included in the performance measures plotted in Figures 1 through 12. When the model is maximally complex, this measure indicates the quality of the feature set, as discussed above. As soon as the complexity of the model is reduced, the model form is no longer an exact expression of the distribution apparent in the test set; assumptions of conditional independence have been introduced into the model. The process used to reduce model complexity assures that each time an assumption of conditional independence is made (i.e., an interdependency between two variables is removed), it is, at least in a local sense, the most appropriate one to have selected based on an analysis of the training data. In Figures 1 through 12 we see that, up to a point, judicious selection of the conditional independence assumptions allows us to reduce model complexity without impacting our ability to characterize the distribution of tags in the test set (i.e., starting from the right, the parametric form remains flat for some time as complexity is decreased). But, in all cases, as the process of reducing model complexity continues, the model loses its ability to properly characterize this distribution. This failure to properly characterize the test set occurs when the interdependencies removed from the model are important in describing the conditional distribution of the tags given the values of the non-classification variables (equation 4). The exact point at which this occurs varies in Figures 1 through 12, but the fact that it does occur is apparent in the drop-off of the measure of form as well as in the increase in misclassification error that accompanies that drop. In all figures, as the measure of form drops, the gap between recall and overall model performance increases, indicating the contribution that the inappropriateness of model form makes to misclassification error.

### As the complexity of the model is reduced, the quality of the parameter estimates improves.

The final factor contributing to misclassification error is the quality of the parameter estimates. The gap between the measure of form and the overall model performance is the error that results from using parameter estimates made from the training data as opposed to using parameters that exactly describe the characteristics of the test set (recall that the only difference between these measures is whether the parameters are estimated from the test set or from the training data). In all figures, this gap shrinks dramatically as the complexity of the model is reduced. The decrease in this gap indicates that the quality of the parameter estimates made from the training data improves as model complexity is reduced. Similarly, this improvement is reflected in recall, which also increases as the complexity of the model is reduced.

### The quality of the non-zero parameter estimates can be isolated.

In the previous subsection, we considered the quality of the parameter estimates by considering the overall model performance. The negative effect of the parameter estimates on this measure includes both losses due to lack of recall and losses due to incorrect tagging. We can isolate the losses due to incorrect tagging in certain cases, namely when the measure of form is 100%. When the measure of form is 100%, there is no error due to the parametric form or to the feature set (see the discussion of the measure of feature-set above). Thus, the lack of precision (i.e., the misclassification error) is due only to the inappropriateness of the parameter estimates for the test set. For four of the words—"bill", "chief", "concern", and "include"—the measure of form for the most complex models is 100%. For these models, the precision is very good, ranging from roughly 95% for "bill" to 100% for "include." What lack of precision there is (for models with measure of form of 100%) is due to inappropriateness of non-zero parameter estimates.

## Discussion

Before concluding, it is important to discuss the interdependencies that exist among the three determinants of model performance. The ideal model is, of course, one in which all three are optimal. But is it possible to design a model that is optimal in all three using a fixed amount of training data? Not surprisingly, the answer for most interesting problems is no. An optimal set of features is one that serves to fully distinguish among the tags being assigned. An optimal set (if one exists) or even a reasonably good set is likely to be large for any interesting problem. Defining a good model of the joint distribution of a large set of variables using a fixed amount of training data is a process of finding the level of model complexity that provides the right balance between quality of form and quality of parameter estimates (where only the most important interdependencies are included at each complexity level).

The need for this balance is demonstrated in Figures 1 through 12 and can be explained as follows. Reducing the complexity of a model entails reducing the number of interdependencies specified in the form of that model and this, in turn, results in a reduction in the number of model parameters. While reducing the number of model parameters increases the quality of the parameter estimates, reducing the number of interdependencies specified in the model results in a loss of information. This loss negatively affects the characterization of the joint distribution by the parametric form. Thus, the best overall model performance is obtained when the appropriate balance is reached.

## Conclusions

This paper described measures for evaluating the three determinants of how well a probabilistic classifier performs on a given test set. These determinants are the appropriateness, for the test set, of the results of (1) features selection, (2) formulation of the parametric form of the model, and (3) parameter estimation. These are part of any model formulation procedure, even if not broken out as separate steps, so the tradeoffs explored in this paper are relevant to a wide variety of methods. The measures were demonstrated in a large experiment, in which they were used to analyze the results of roughly 300 classifiers that perform word-sense disambiguation. These evaluations suggest that the three determinants of model performance are not independent and that the best overall model performance is obtained when they are appropriately balanced.

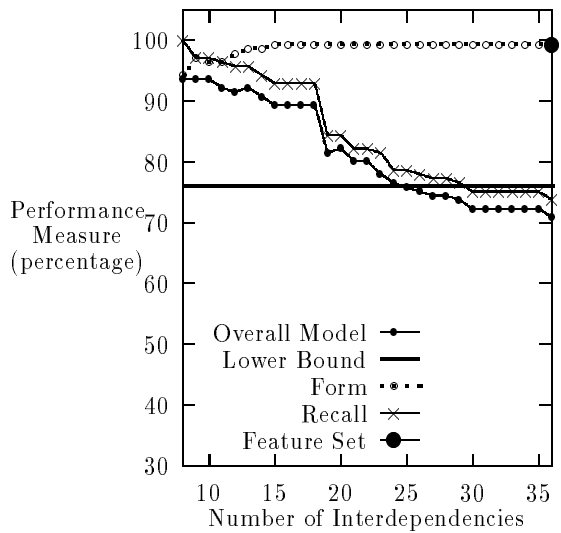

Figure 1: "agree"

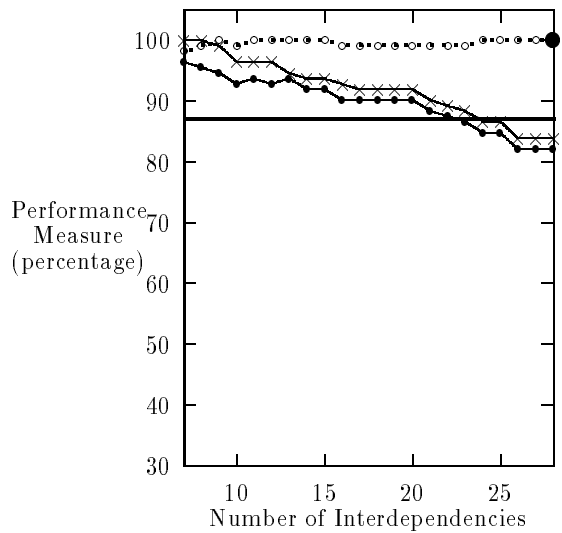

Figure 3: "chief"

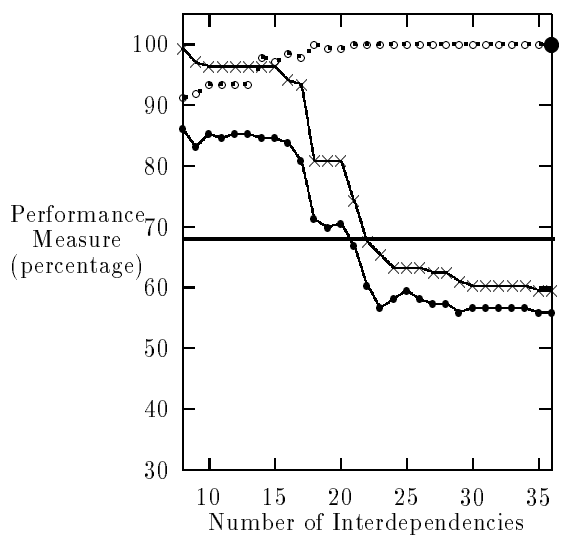

Figure 2: "bill"

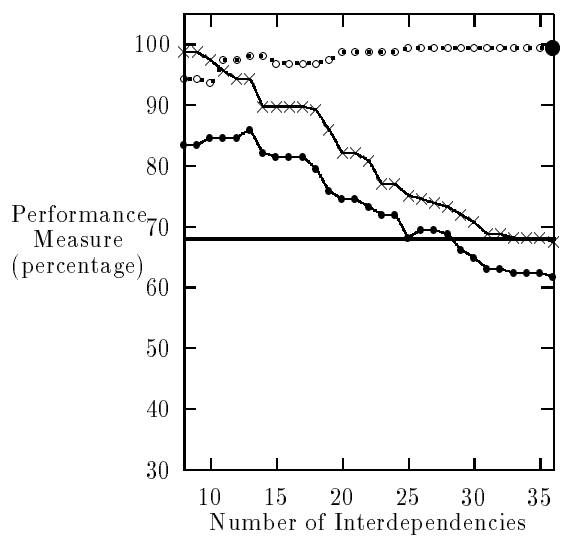

Figure 4: "close"

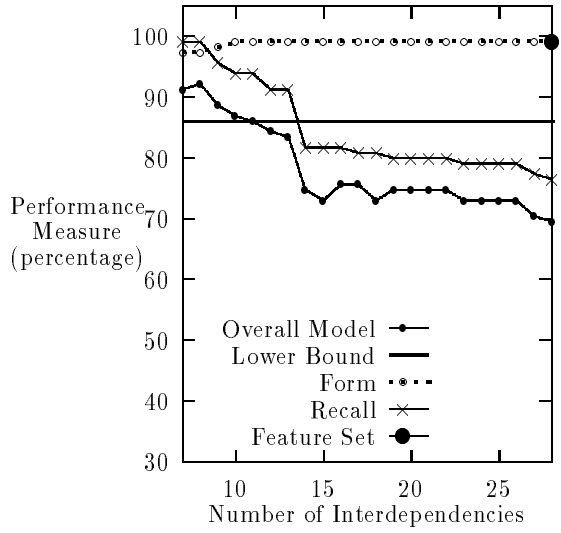

Figure 5: "common"

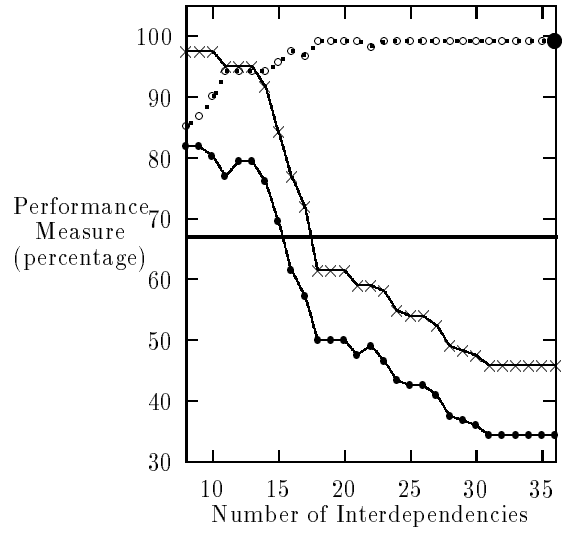

Figure 7: "drug"

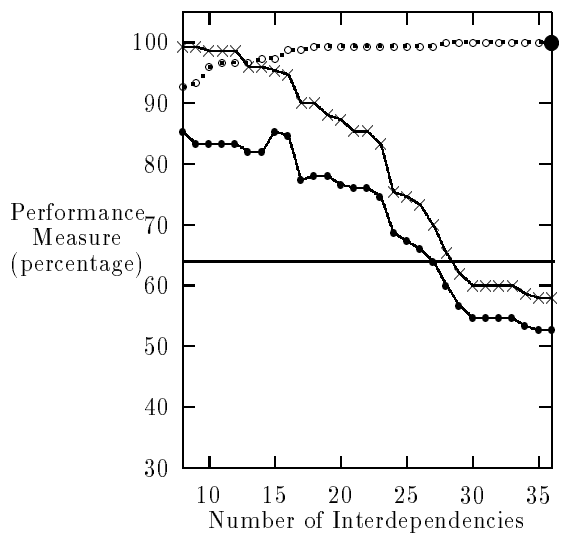

Figure 6: "concern"

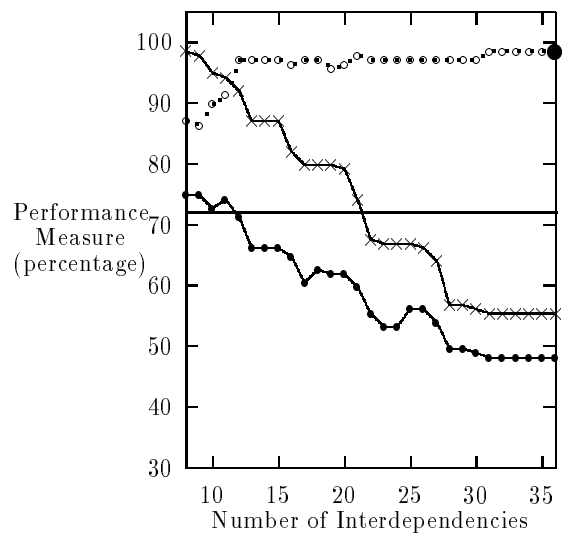

Figure 8: "help"

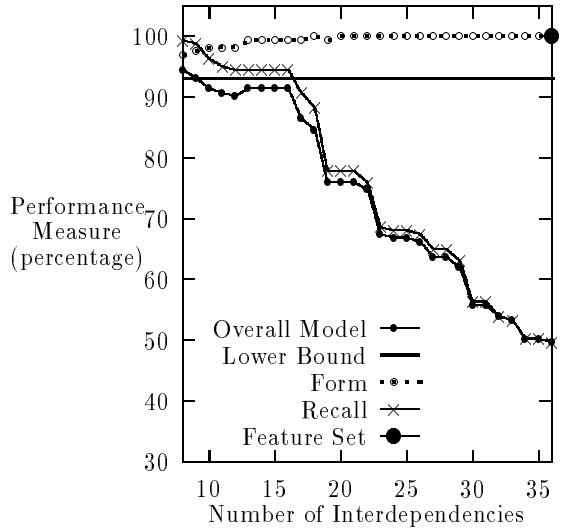

Figure 9: "include"

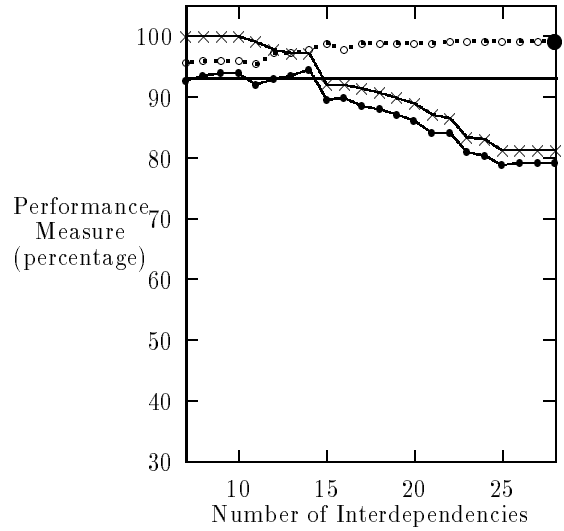

Figure 11: "last"

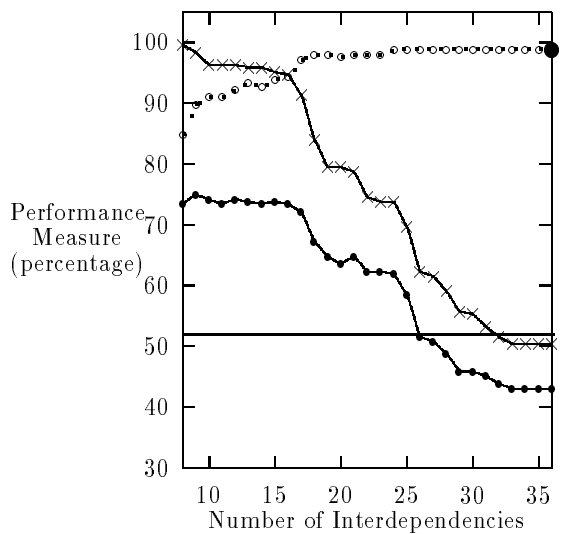

Figure 10: "interest"

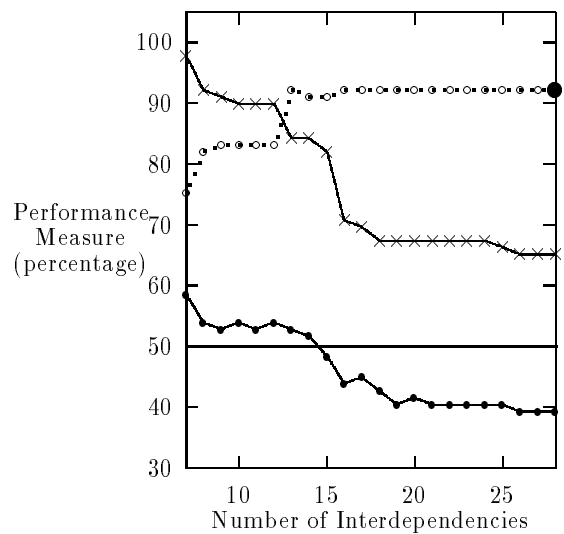

Figure 12: "public"